\documentclass[10pt,twocolumn,letterpaper]{article}

\usepackage{iccv}
\usepackage{times}
\usepackage{epsfig}
\usepackage{graphicx}
\usepackage{amsmath,amssymb}
\usepackage{commath}
\usepackage{enumitem}
\usepackage{times}
\usepackage{textcomp}

\usepackage{subcaption}
\usepackage{colortbl} 
\usepackage{xcolor}
\usepackage{soul}

\usepackage[pagebackref=true,breaklinks=true,letterpaper=true,colorlinks,bookmarks=false]{hyperref}
\usepackage{url}

\iccvfinalcopy 

\def\iccvPaperID{560} 

\newcommand{\ve}[1]{\mathbf{#1}} 

\ificcvfinal\fi
\begin{document}
	
	\title{How far are we from solving the 2D \& 3D Face Alignment problem? (and a dataset of 230,000 3D facial landmarks)}
	
	\author{Adrian Bulat  and 
		Georgios Tzimiropoulos\\
		Computer Vision Laboratory, The University of Nottingham\\
		Nottingham, United Kingdom\\
		{\tt\small \{adrian.bulat, yorgos.tzimiropoulos\}@nottingham.ac.uk}}
	
	\maketitle

	\begin{abstract}
		This paper investigates how far a very deep neural network is from attaining close to saturating performance on existing 2D and 3D face alignment datasets. To this end, we make the following 5 contributions: (a) we construct, for the first time, a very strong baseline by combining a state-of-the-art architecture for landmark localization with a state-of-the-art residual block, train it on a very large yet synthetically expanded 2D facial landmark dataset and finally evaluate it on \textit{all other} 2D facial landmark datasets. (b) We create a guided by 2D landmarks network which converts 2D landmark annotations to 3D and unifies all existing datasets, leading to the creation of LS3D-W, the largest and most challenging 3D facial landmark dataset to date (\texttildelow230,000 images). (c) Following that, we train a neural network for 3D face alignment and evaluate it on the newly introduced LS3D-W. (d) We further look into the effect of all ``traditional''  factors affecting face alignment performance like large pose, initialization and resolution, and  introduce a ``new'' one, namely the size of the network. (e) We show that both 2D and 3D face alignment networks achieve performance of remarkable accuracy which is probably close to saturating the datasets used. Training and testing code as well as the dataset can be downloaded from \url{https://www.adrianbulat.com/face-alignment/} 
		
	\end{abstract}
	
	\section{Introduction}
	With the advent of Deep Learning and the development of large annotated datasets, recent work has shown results of unprecedented accuracy even on the most challenging computer vision tasks.  In this work, we focus on landmark localization, in particular, on facial landmark localization, also known as face alignment, arguably one of the most heavily researched topics in computer vision over the last decades. Very recent work on landmark localization using Convolutional Neural Networks (CNNs) has pushed the boundaries in other domains like human pose estimation \cite{toshev2014deeppose, tompson2014joint, pfister2015flowing, insafutdinov2016deepercut, pishchulin2016deepcut, wei2016convolutional, newell2016stacked, bulat2016human}, yet it remains unclear what has been achieved so far for the case of face alignment. The aim of this work is to address this gap in literature.
	
	Historically, different techniques have been used for landmark localization depending on the task in hand. For example, work in human pose estimation, prior to the advent of neural networks, was primarily based on pictorial structures \cite{felzenszwalb2005pictorial} and sophisticated extensions \cite{yang2011articulated, pishchulin2013poselet, tian2012exploring, sapp2013modec, pishchulin2013strong} due to their ability to model large appearance changes and accommodate a wide spectrum of human poses. Such methods though have not been shown capable of achieving the high degree of accuracy exhibited by cascaded regression methods for the task of face alignment \cite{dollar2010cascaded, Cao2012shaperegression,xiong2013supervised,zhu2016face,tzimiropoulos2015project}. On the other hand, the performance of cascaded regression methods is known to deteriorate for cases of inaccurate initialisation, and large (and unfamiliar) facial poses when there is a significant number of self-occluded landmarks or large in-plane rotations. 
	
	More recently, fully Convolutional Neural Network architectures based on heatmap regression have revolutionized human pose estimation \cite{toshev2014deeppose, tompson2014joint, pfister2015flowing, insafutdinov2016deepercut, pishchulin2016deepcut, wei2016convolutional, newell2016stacked, bulat2016human} producing results of remarkable accuracy even for the most challenging  datasets \cite{andriluka20142d}. Thanks to their end-to-end training and little need for hand engineering, such methods can be readily applied to the problem of face alignment. Following this path, our main contribution is to construct and train such a powerful network for face alignment and investigate for the first time how far it is from attaining close to saturating performance on all existing 2D face alignment datasets and a newly introduced large scale 3D dataset. More specifically, \textbf{our contributions} are:
	\begin{figure*}[!htb]
		\begin{center}
			\centering
			\includegraphics[height=1.4in,trim={0.5cm 0.5cm 0.5cm 0.5cm},clip]{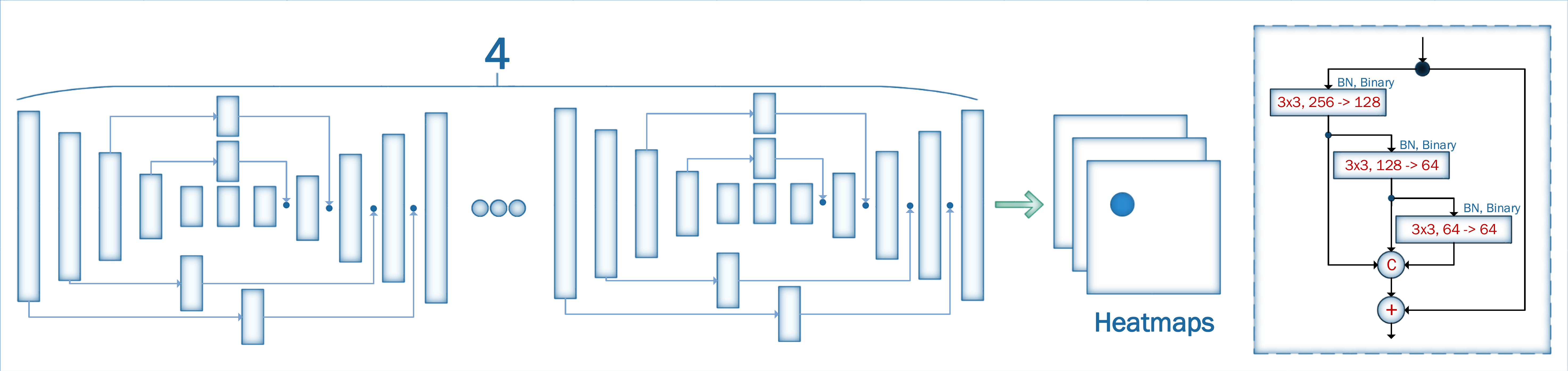}
		\end{center}
		\caption{The Face Alignment Network (FAN) constructed by stacking four HGs in which all bottleneck blocks (depicted as rectangles) were replaced with the hierarchical, parallel and multi-scale block of \cite{bulat2017binarized}.}
		\label{fig:block}
	\end{figure*} 
	\vspace{-0.05cm}
	\begin{enumerate}[leftmargin=*]
		\setlength\itemsep{-0.35em}
		\item
		We construct, for the first time, a very strong baseline by combining a state-of-the-art architecture for landmark localization with a state-of-the-art residual block and train it on a very large yet synthetically expanded 2D facial landmark dataset. Then, we evaluate it on \textit{all other} 2D datasets (\texttildelow230,000 images), investigating how far are we from solving 2D face alignment.
		\item
		In order to overcome the scarcity of 3D face alignment datasets, we further propose a guided-by-2D landmarks CNN which converts 2D annotations to 3D \footnote{The 3D annotations are actually the 2D projections of the 3D facial landmarks but for simplicity we will just call them 3D. In the supplementary material, we present a method for extending them to full 3D.} and use it to create LS3D-W, the largest and most challenging 3D facial landmark dataset to date (\texttildelow 230,000 images), obtained from unifying almost all existing datasets to date.
		\item
		Following that, we train a 3D face alignment network and then evaluate it on the newly introduced large scale 3D facial landmark dataset, investigating how far are we from solving 3D face alignment.  
		\item
		We further look into the effect of all ``traditional''  factors affecting face alignment performance like large pose, initialization and resolution, and  introduce a ``new'' one, namely the size of the network. 
		\item
		We show that both 2D and 3D face alignment networks achieve performance of remarkable accuracy which is probably close to saturating the datasets used.      
	\end{enumerate}
	
	
	\section{Closely related work}
	This Section reviews related work on face alignment and landmark localization. Datasets are described in detail in the next Section.\newline 
	\textbf{2D face alignment.} Prior to the advent of Deep Learning, methods based on cascaded regression had emerged as the state-of-the-art in 2D face alignment, see for example \cite{Cao2012shaperegression,xiong2013supervised,zhu2016face,tzimiropoulos2015project}. Such methods are now considered to have largely ``solved'' the 2D face alignment problem for faces with controlled pose variation like the ones of LFPW \cite{belhumeur2011localizing}, Helen \cite{le2012interactive} and 300-W \cite{sagonas2013semi}. 
	
	We will keep the main result from these works, namely their performance on the frontal dataset of LFPW \cite{belhumeur2011localizing}. This performance will be used as a measure of comparison of how well the methods described in this paper perform assuming that a method achieving a similar error curve on a different dataset is close to saturating that dataset. \newline 
	\textbf{CNNs for face alignment.} By no means we are the first to use CNNs for face alignment. The method of \cite{sun2013deep} uses a CNN cascade to regress the facial landmark locations. The work in \cite{zhang2014facial} proposes multi-task learning for joint facial landmark localization and attribute classification. More recently, the method of \cite{trigeorgismnemonic} extends \cite{xiong2013supervised} within recurrent neural networks. All these methods have been mainly shown effective for the near-frontal faces of 300-W \cite{sagonas2013semi}.
	
	
	Recent works on large pose and 3D face alignment includes  \cite{jourabloo2016large, zhu2016face} which perform face alignment by fitting a 3D Morphable Model (3DMM) to a 2D facial image. The work in \cite{jourabloo2016large} proposes to fit a dense 3DMM using a cascade of CNNs. The approach of \cite{zhu2016face} fits a 3DMM in an iterative manner through a single CNN which is augmented by additional input channels (besides RGB) representing shape features at each iteration. More recent works that are closer to the methods presented in this paper are \cite{bulat2016convolutional} and \cite{bulat2016two}. Nevertheless, \cite{bulat2016convolutional} is evaluated on \cite{jourabloo2016large} which is a relatively small dataset (3900 images for training and 1200 for testing) and \cite{bulat2016two} on \cite{jeni2016first} which is of moderate size (16,2000 images for training and 4,900 for testing), includes mainly images collected in the lab and does not cover the full spectrum of facial poses. Hence, the results of \cite{bulat2016convolutional} and \cite{bulat2016two} are not conclusive in regards to the main questions posed in our paper. \newline 
	\textbf{Landmark localization.} A detailed review of state-of-the-art methods on landmark localization for human pose estimation is beyond the scope of this work, please see \cite{toshev2014deeppose, tompson2014joint, pfister2015flowing, insafutdinov2016deepercut, pishchulin2016deepcut, wei2016convolutional, newell2016stacked, bulat2016human}. For the needs of this work, we built a powerful CNN for 2D and 3D face alignment based on two components: (a) the state-of-the-art Hour-Glass (HG) network of \cite{newell2016stacked}, and (b) the hierarchical, parallel \& multi-scale block recently proposed in \cite{bulat2017binarized}. In particular, we replaced the bottleneck block \cite{he2016identity} used in \cite{newell2016stacked} with the block proposed in \cite{bulat2017binarized}. \newline 
	\textbf{Transferring landmark annotations.} There are a few works that have attempted to unify facial alignment datasets by transferring landmark annotations, typically through exploiting common landmarks across datasets \cite{zhu2014transferring, smith2014collaborative, zhang2015leveraging}. Such methods have been primarily shown to be successful when landmarks are transferred from more challenging to less challenging images, for example in \cite{zhu2014transferring} the target dataset is LFW \cite{huang2007labeled} or \cite{smith2014collaborative} provides annotations only for the relatively easy images of AFLW \cite{kostinger2011annotated}. Hence, the community primarily relies on the unification performed manually by the 300-W challenge \cite{sagonas2013300} which contains less than 5,000 near frontal images annotated from a 2D perspective. 
	
	Using 300-W-LP \cite{zhu2016face} as a basis, this paper presents the first attempt to provide 3D annotations for \textit{all other} datasets, namely AFLW-2000 \cite{zhu2016face} (2,000 images), 300-W test set \cite{sagonas2016300} (600 images), 300-VW \cite{shen2015first} (218,595 frames), and Menpo training set (9,000 images). To this end, we propose a guided-by-2D landmarks CNN which converts 2D annotations to 3D and unifies all aforementioned datasets.

	
	\section{Datasets}
	
	In this Section, we provide a description of how existing 2D and 3D datasets were used for training and testing for the purposes of our experiments. We note that the 3D annotations preserve correspondence across pose as opposed to the 2D ones and, in general, they should be preferred. We emphasize that the 3D annotations are actually the 2D projections of the 3D facial landmark coordinates but for simplicity we will just call them 3D. In the supplementary material, we present a method for extending these annotations to full 3D. Finally, we emphasize that we performed cross-database experiments only. 
	
	\begin{table}[!htbp]
		\small
		\begin{center}
			\begin{tabular}{|l|c|c|c|c|c|}
				\hline
				Dataset &  Size & pose & annot. & synt.\\
				\hline\hline
				300-W & 4,000 &[$-45^o$, $45^o$]  &2D & No\\
				300W-LP-2D & 61,225 & [$-90^o$, $90^o$] & 2D & Yes\\
				300W-LP-3D  & 61,225 & [$-90^o$, $90^o$]& 3D & Yes\\
				AFLW2000-3D  & 2,000 & [$-90^o$, $90^o$] & 3D & No\\
				300-VW & 218,595 & [$-45^o$, $45^o$] & 2D & No\\
				\textbf{LS3D-W (ours)} & 230,000 & [$-90^o$, $90^o$] & 3D & No\\
				\hline
			\end{tabular}
		\end{center}
		\caption{Summary of the most popular face alignment datasets and their main characteristics.}
		\label{tab:all_datasets}
	\end{table}
	
	\subsection{Training datasets}
	For training and validation, we used 300-W-LP \cite{zhu2016face}, a synthetically expanded version of 300-W \cite{sagonas2013300}. 300-W-LP provides both 2D and 3D landmarks allowing for training models and conducting experiments using both types of annotations. For some 2D experiments, we also used the original 300-W dataset \cite{sagonas2013300} for fine tuning, only. This is because the 2D landmarks of 300-W-LP are not entirely compatible with the 2D landmarks of the test sets used in our experiments (i.e. 300-W test set, \cite{sagonas2016300}, 300-VW \cite{shen2015first} and Menpo \cite{menpo}), but the original annotations from 300-W are. 
	\newline \textbf{300-W.} 300-W \cite{sagonas2013300} is currently the most widely-used \textit{in-the-wild} dataset for 2D face alignment. The dataset itself is a concatenation of a series of smaller datasets: LFPW \cite{belhumeur2013localizing}, HELEN \cite{le2012interactive}, AFW \cite{zhu2012face} and iBUG \cite{sagonas2013semi}, where each image was re-annotated in a consistent manner using the 68 2D landmark configuration of Multi-PIE \cite{gross2010multi}. The dataset contains in total \texttildelow 4,000 near frontal facial images. \newline
	\textbf{300W-LP-2D and 300W-LP-3D.} 300-W-LP is a synthetically generated dataset obtained by rendering the faces of 300-W into larger poses, ranging from $-90^0$ to $90^0$, using the profiling method of \cite{zhu2016face}. The dataset contains 61,225 images providing both 2D (300W-LP-2D) and 3D landmark annotations (300W-LP-3D). 
	
	\subsection{Test datasets}
	
	This Section describes the test sets used for our 2D and 3D experiments. Observe that there is a large number of 2D datasets/annotations which are however  problematic for moderately large poses (2D landmarks lose correspondence) and that the only in-the-wild 3D test set is AFLW2000-3D \cite{zhu2016face} \footnote{The data from \cite{jeni2016first} includes mainly images collected in the lab and do not cover the full spectrum of facial poses.}. We address this significant gap in 3D face alignment datasets in Section~\ref{sec:large_scale}. 
	
	\subsubsection{2D datasets}
	
	\textbf{300-W test set.} The 300-W test set consists of the 600 images used for the evaluation purposes of the 300-W Challenge \cite{sagonas2016300}. The images are split in two categories: \textit{Indoor} and \textit{Outdoor}. All images were annotated with the same 68 2D landmarks as the ones used in the 300-W data set. \newline
	\textbf{300-VW.} 300-VW\cite{shen2015first} is a large-scale face tracking dataset, containing 114 videos and in total 218,595 frames. From the total of 114 videos, 64 are used for testing and 50 for training.  The test videos are further separated into three categories (A, B, and C) with the last one being the most challenging. It is worth noting that some videos (especially from category C) contain very low resolution/poor quality faces. Due to the semi-automatic annotation approach (see \cite{shen2015first} for more details), in some cases, the annotations for these videos are not so accurate (see Fig.~\ref{fig:300VW_visual}). Another source of annotation error is caused by facial pose, i.e. large poses are also not accurately annotated (see Fig.~\ref{fig:300VW_visual}). \newline
	\textbf{Menpo.} Menpo is a recently introduced dataset \cite{menpo} containing landmark annotations for about 9,000 faces from FDDB \cite{jain2010fddb} and ALFW. Frontal faces were annotated in terms of 68 landmarks using the same annotation policy as the one of 300-W but profile faces in terms of 39 different landmarks which are not in correspondence with the landmarks from the 68-point mark-up. 
	
	\subsubsection{3D datasets}
	
	\textbf{AFLW2000-3D.} AFLW2000-3D \cite{zhu2016face} is a dataset constructed by re-annotating the first 2000 images from AFLW \cite{kostinger2011annotated} using 68 3D landmarks in a consistent manner with the ones from 300W-LP-3D. The faces of this dataset contain large-pose variations (yaw from $-90^o$ to $90^o$), with various expressions and illumination conditions. However, some annotations, especially for larger poses or occluded faces are not so accurate (see Fig.~\ref{fig:large_scale_aflw_2000}).

	
	
	
	
	\subsection{Metrics} 
	
	Traditionally, the metric used for face alignment is the point-to-point Euclidean distance normalized by the interocular distance  \cite{cristinacce2006feature, sagonas2013300, shen2015first}. However, as noted in \cite{zhu2012face}, this error metric is biased for profile faces for which the interocular distance can be very small. Hence, we normalize by the bounding box size. In particular, we used the Normalized Mean Error defined as: 
	
	\begin{equation}
	\textrm{NME} = \frac{1}{N} \sum_{k=1}^{N} \frac{\norm{ \ve{x}_k-\ve{y}_{k} }_{2} }{d}, \label{eq:additive0}
	\end{equation}
	
	\noindent where $\ve{x}$ denotes the ground truth landmarks for a given face, $\ve{y}$ the corresponding prediction and $d$ is the square-root of the ground truth bounding box, computed as $d = \sqrt{w_{bbox}*h_{bbox}}$. Although we conducted both 2D and 3D experiments, we opted to use the same bounding box definition for both experiments; in particular we used the bounding box calculated from the 2D landmarks. This way, we can readily compare the accuracy achieved in 2D and 3D. 
	
	

	
	\section{Method}
	This Section describes FAN, the network used for 2D and 3D face alignment. It also describes 2D-to-3D FAN, the network used for constructing the very large scale 3D face alignment dataset (LS3D-W) containing more than 230,000 3D landmark annotations.
	
	\subsection{2D and 3D Face Alignment Networks} 
	
	We coin the network used for our experiments simply Face Alignment Network (FAN). To our knowledge, it is the first time that such a powerful network is trained and evaluated for large scale 2D/3D face alignment experiments. 
	
	We construct FAN based on one of the state-of-the-art architectures for human pose estimation, namely the \textit{Hour-Glass} (HG) network of \cite{newell2016stacked}. In particularly, we used a stack of four HG networks (see Fig.~\ref{fig:block}). While \cite{newell2016stacked} uses the bottleneck block of \cite{he2016deep} as the main building block for the HG, we go one step further and replace the bottleneck block with the recently introduced hierarchical, parallel and multi-scale block of \cite{bulat2017binarized}. As it was shown in \cite{bulat2017binarized}, this block outperforms the original bottleneck of \cite{he2016deep} when the same number of network parameter were used. Finally, we used 300W-LP-2D and 300W-LP-3D to train 2D-FAN and 3D-FAN, respectively. 
	
	\subsection{2D-to-3D Face Alignment Network} \label{sec:2D-3D}
	\begin{figure}[!htb]
		\begin{center}
			\centering
			\includegraphics[height=1.0in,trim={0.5cm 0.5cm 0.5cm 0.5cm},clip]{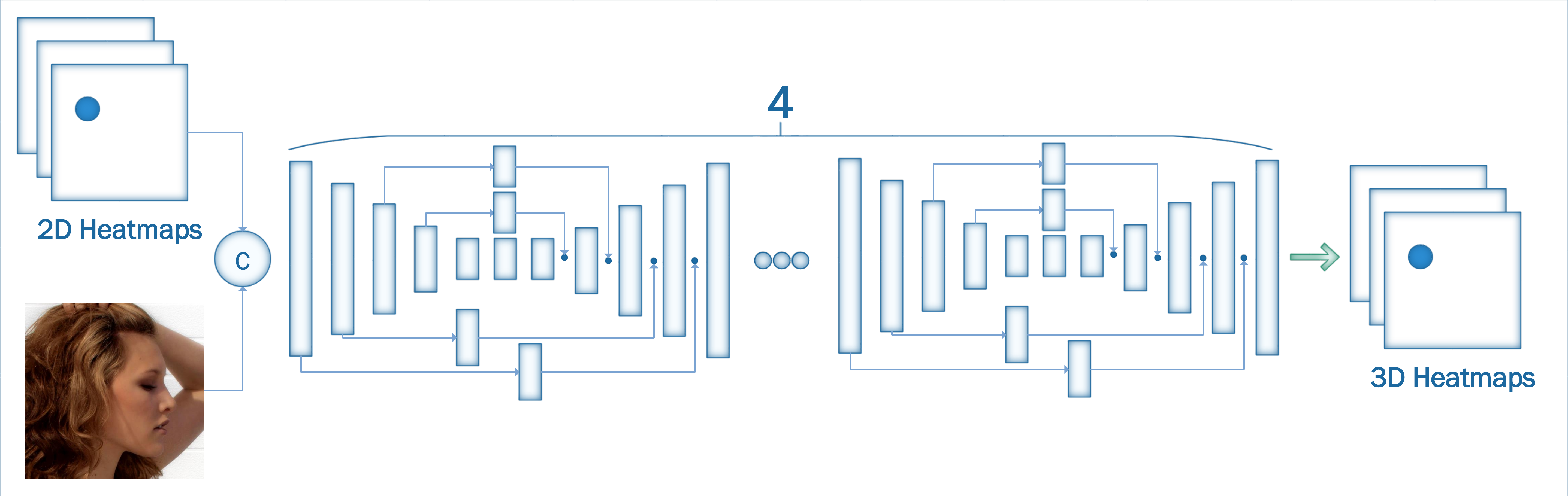}
		\end{center}
		\caption{The 2D-to-3D-FAN network used for the creation of the LS3D-W dataset. The network takes as input the RGB image and the 2D landmarks and outputs the corresponding 2D projections of the 3D landmarks.}
		\label{fig:fan-full}
	\end{figure}
	Our aim is to create the very first very large scale dataset of 3D facial landmarks for which annotations are scarce. To this end, we followed a guided-based approach in which a FAN for predicting 3D landmarks is guided by 2D landmarks. In particular, we created a 3D-FAN in which the input RGB channels have been augmented with 68 additional channels, one for each 2D landmark, containing a 2D Gaussian with std = 1px centered at each landmark's location. We call this network 2D-to-3D FAN. Given the 2D facial landmarks for an image, 2D-to-3D FAN converts them to 3D.  To train 2D-to-3D FAN, we used 300-W-LP which provides both 2D and 3D annotations for the same image. We emphasize again that the 3D annotations are actually the 2D projections of the 3D coordinates but for simplicity we call them 3D. Please see supplementary material for extending these annotations to full 3D. 
	
	
	
	
	\subsection{Training}
	
	
	For all of our experiments, we independently trained three distinct networks: 2D-FAN, 3D-FAN, and 2D-to-3D-FAN. For the first two networks, we set the initial learning rate to $10^{-4}$ and used a minibatch of 10. During the process, we dropped the learning rate to $10^{-5}$ after 15 epochs and to $10^{-6}$ after another 15, training for a total of 40 epochs. We also applied random augmentation: flipping, rotation (from $-50^o$ to $50^o$), color jittering, scale noise (from 0.8 to 1.2) and random occlusion. The 2D-to-3D-FAN model was trained by following a similar procedure increasing the amount of augmentation even further: rotation (from $-70^o$ to $70^o$) and scale (from 0.7 to 1.3). Additionally, the learning rate initially was set to $10^{-3}$. All networks were implemented in Torch7 \cite{collobert2011torch7} and trained using rmsprop \cite{tieleman2012lecture}.
	
	
	\section{2D face alignment} \label{sec:2D}
	
	This Section evaluates 2D-FAN (trained on 300-W-LP-2D), on 300-W test set, 300-VW (both training and test sets), and Menpo (frontal subset). Overall, 2D-FAN is evaluated on more than 220,000 images. Prior to reporting our results, the following points need to be emphasized: 
	\begin{enumerate}[leftmargin=*]
		\setlength\itemsep{-0.35em}
		\item
		300-W-LP-2D contains a wide range of poses (yaw angles in $[-90^\circ, 90^\circ])$, yet it is still a synthetically generated dataset as this wide spectrum of poses were produced by warping the nearly frontal images of the 300-W dataset. It is evident that this lack of real data largely increases the difficulty of the experiment. 
		\item 
		The 2D landmarks of 300-W-LP-2D that 2D-FAN was trained on are slightly different from the 2D landmarks of the 300-W test set, 300-VW and Menpo. To alleviate this, the 2D-FAN was further fine-tuned on the original 300-W training set for a few epochs. Although this seems to resolve the issue, this discrepancy obviously increases the difficulty of the experiment. 
		\item
		We compare the performance of 2D-FAN on all the aforementioned datasets with that of an unconventional baseline: the performance of a recent state-of-the-art method, namely MDM \cite{trigeorgismnemonic} on LFPW test set, initialized with the ground truth bounding boxes. We call this result MDM-on-LFPW.  As there is very little performance progress made on the frontal dataset of LFPW over the past years, we assume that a state-of-the-art method like MDM (nearly) saturates it. Hence, we use the produced error curve to compare how well our method does on the much more challenging aforementioned test sets.
	\end{enumerate}
	
	\begin{figure}[!htb]
		\begin{center}
			\centering
			\includegraphics[height=1.67in,trim={0.5cm 0.5cm 0.5cm 0.5cm},clip]{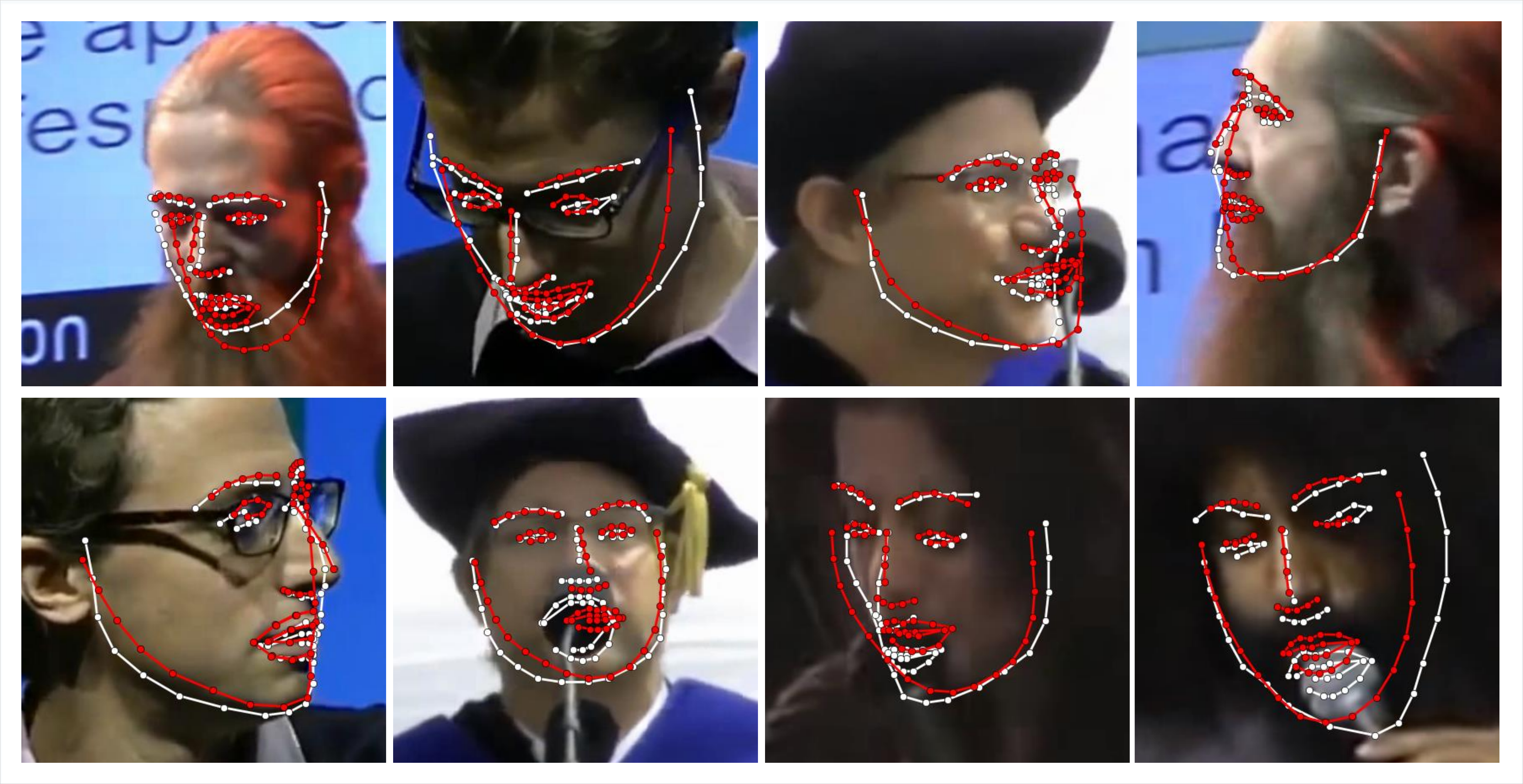}
		\end{center}
		\caption{Fittings with the highest error from 300-VW (NME 6.8-7\%). Red: ground truth. White: our predictions. In most cases, our predictions are more accurate than the ground truth.}
		\label{fig:300VW_visual}
	\end{figure}
	
\begin{figure*}[!htb]
		\vspace{-0.5cm}
		\centering
		\begin{subfigure}[t]{0.3\textwidth}
			\includegraphics[height=1.7in,trim={0cm 0cm 0cm 0cm},clip]{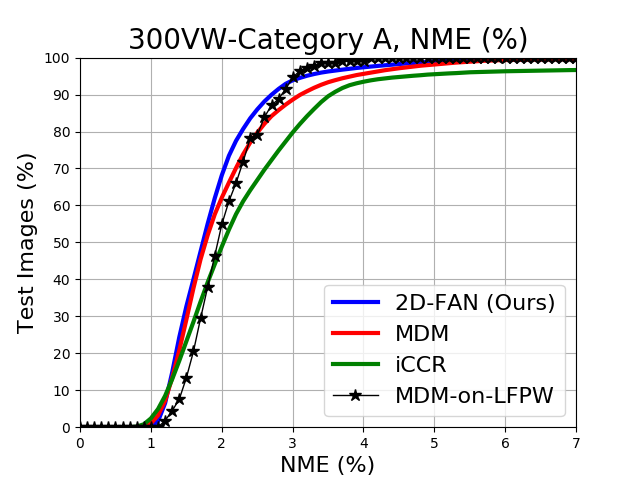}
			\caption{300VW-Category A}
			\label{fig:300VW-A}
		\end{subfigure}
		~
		\begin{subfigure}[t]{0.3\textwidth}
			\includegraphics[height=1.7in,trim={0cm 0cm 0cm 0cm},clip]{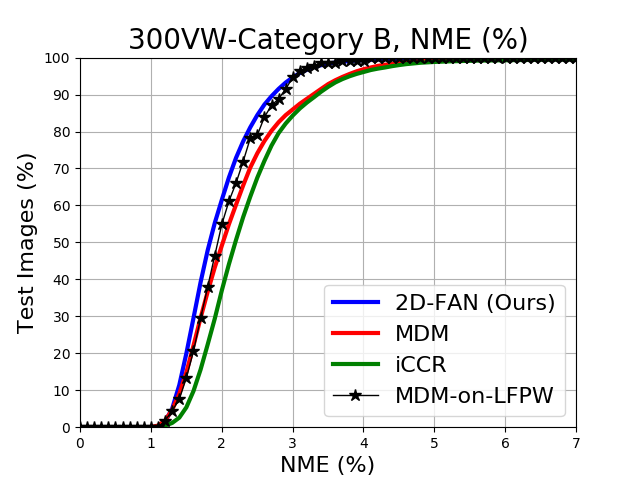}
			\caption{300VW-Category B}
			\label{fig:300VW-B}
		\end{subfigure}
		~
		\begin{subfigure}[t]{0.3\textwidth}
			\includegraphics[height=1.7in,trim={0cm 0cm 0cm 0cm},clip]{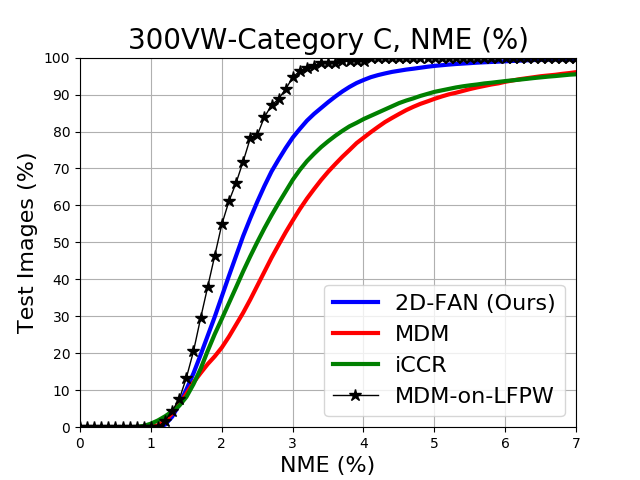}
			\caption{300VW-Category C}
			\label{fig:300VW-C}
		\end{subfigure}
		~
		\begin{subfigure}[t]{0.45\textwidth}
			\centering
			\includegraphics[height=1.7in,trim={0cm 0cm 0cm 0cm},clip]{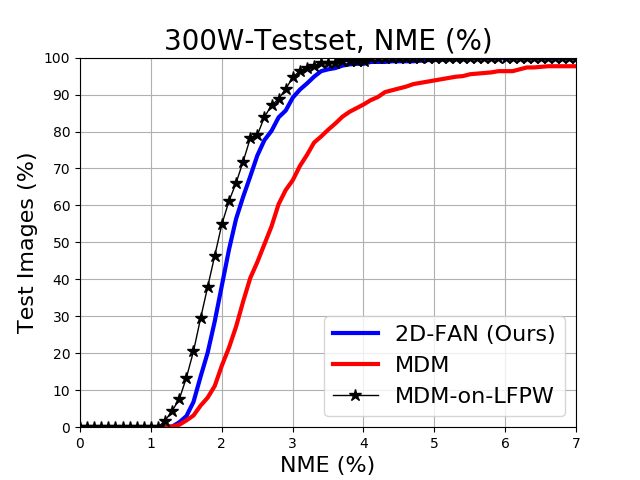}
			\caption{300-W Testset (Indoor and Outdoor subset).}
			\label{fig:300W_bb}
		\end{subfigure}
		~
		\begin{subfigure}[t]{0.45\textwidth}
			\centering
			\includegraphics[height=1.7in,trim={0cm 0cm 0cm 0cm},clip]{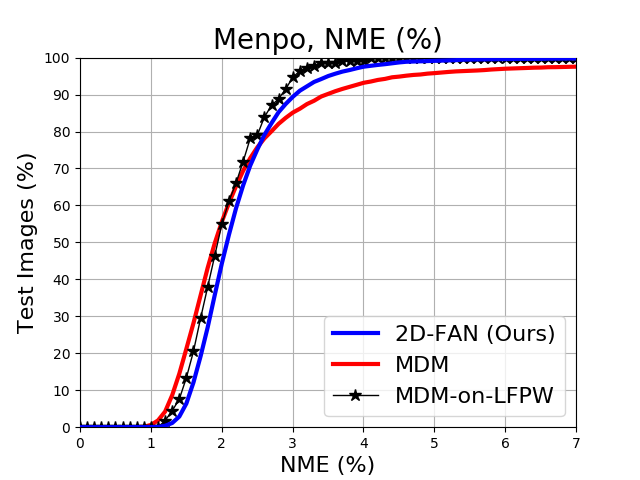}
			\caption{Menpo (on faces annotated with 68 points).}
			\label{fig:menpo-2D_bb}  
		\end{subfigure}
		
		\caption{2D face alignment experiments: NME (all 68 points used) on 300-VW (a-c), 300-W Testset (d) and Menpo (e). Our model is called 2D-FAN. MDM is initialized with ground truth bounding boxes.
			\textbf{Note: MDM-on-LFPW is not a method but the curve produced by running MDM on LFPW test set, initialized with the ground truth bounding boxes.}}
		
		\label{fig:results_2D_all}
	\end{figure*}

	
	The cumulative error curves for our 2D experiments on 300-VW, 300-W test set and Menpo are shown in Fig.~\ref{fig:results_2D_all}. We additionally report the performance of MDM on all datasets initialized by ground truth bounding boxes, ICCR, the state-of-the-art face tracker of \cite{sanchez2016cascaded}, on 300-VW (the only tracking dataset), and our unconventional baseline (called MDM-on-LFPW). Comparison with a number of methods in terms of AUC are also provided in Table \ref{tab:method_2d_auc}.
	
	\begin{table*}[!htbp]
		\small
		\begin{center}
			\begin{tabular}{|l|c|c|c|c|c|}
				\hline
				Dataset &  \textbf{2D-FAN(Ours)} & MDM\cite{trigeorgismnemonic} & iCCR\cite{sanchez2016cascaded} & TCDCN\cite{zhang2014facial} & CFSS\cite{zhu2015cfss}\\
				\hline\hline
				300VW-A & \textbf{72.1\%} &70.2 \% &65.9\% & - & - \\
				300VW-B  & \textbf{71.2\%} & 67.9 \%& 65.5\% & - & - \\
				300VW-C  & \textbf{64.1\%} & 54.6\%& 58.1\% & - & - \\
				Menpo  & \textbf{67.5\%} & 67.1\% & - & 47.9\% & 60.5\% \\
				300W  & \textbf{66.9\%} & 58.1\% & - & 41.7\% & 55.9\% \\
				\hline
			\end{tabular}
		\end{center}
		\caption{AUC (calculated for a threshold of 7\%) on all major 2D face alignment datasets. MDM, CFSS and TCDCN were evaluated using ground truth bounding boxes and the openly available code.}
		\label{tab:method_2d_auc}
	\end{table*}
	
\begin{figure}[!htb]
		\begin{center}
			\centering
			\includegraphics[height=1.67in,trim={0.5cm 0.5cm 0.5cm 0.5cm},clip]{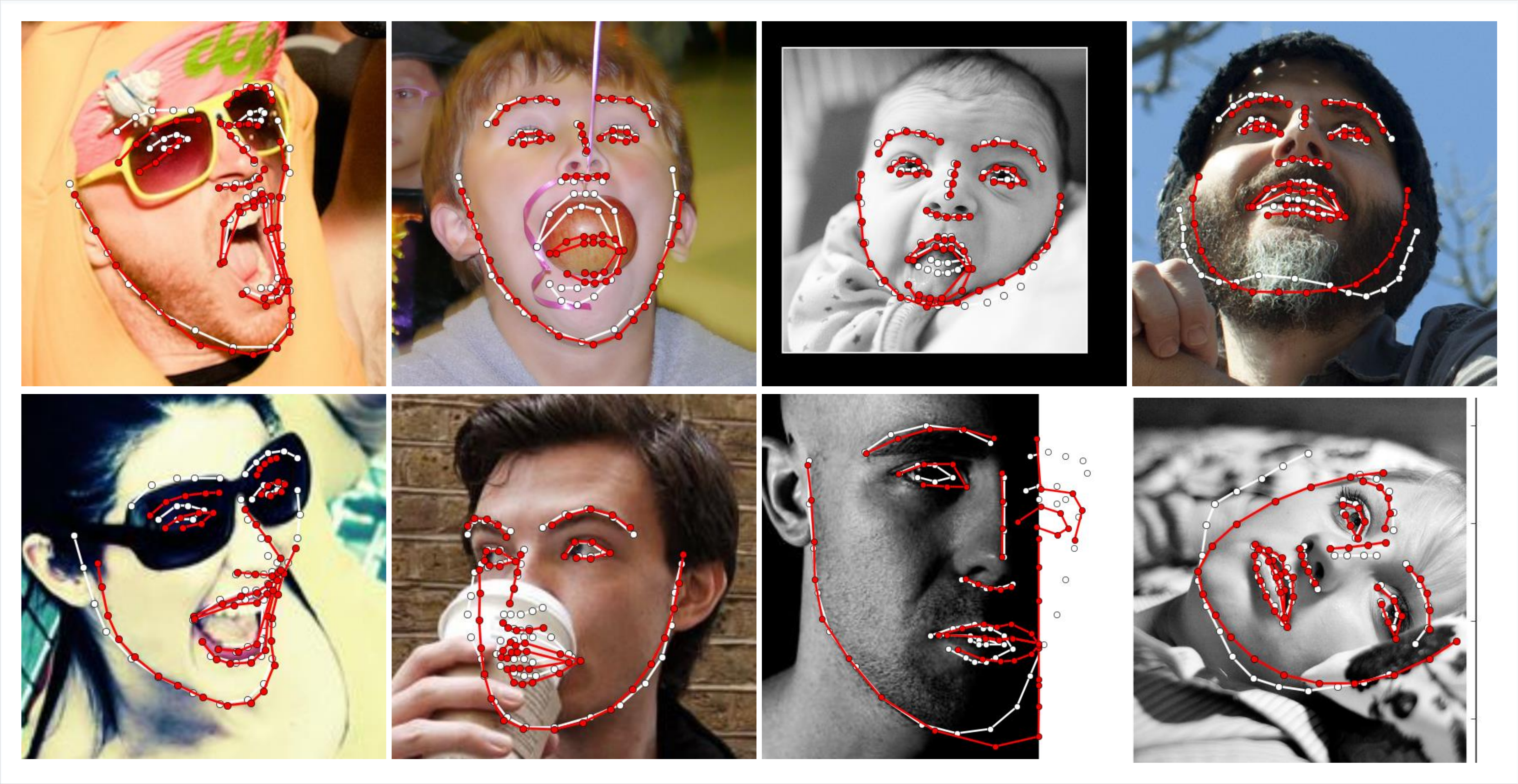}
		\end{center}
		\caption{ Fittings with the highest error from 300-W test set (first row) and Menpo (second row) (NME 6.5-7\%). Red: ground truth. White: our predictions. In most cases, our predictions are more accurate than the ground truth.}
		\label{fig:menpo_300_visual}
	\end{figure}

	
	With the exception of Category C of 300-VW, it is evident that 2D-FAN achieves literally the same performance on all datasets, outperforming MDM and ICCR, and, notably, matching the performance of MDM-on-LFPW. Out of 7,200 images (from Menpo and 300-W test set), there are in total only 18 failure cases, which represent 0.25\% of the images (we consider a failure a fitting with NME $>$ 7\%). After removing these cases, the 8 fittings with the highest error for each dataset are shown in Fig.~\ref{fig:menpo_300_visual}.  
	
	Regarding the Category C of 300-VW, we found that the main reason for this performance drop is the quality of the annotations which were obtained in a semi-automatic manner. After removing all failure cases (101 frames representing 0.38\% of the total number of frames), Fig.~\ref{fig:300VW_visual} shows the quality of our predictions vs the ground truth landmarks for the 8 fittings with the highest error for this dataset. It is evident that in most cases our predictions are more accurate. \newline  
	\textbf{Conclusion:} Given that 2D-FAN matches the performance of MDM-on-LFPW, we conclude that 2D-FAN achieves near saturating performance on the above 2D datasets. Notably, this result was obtained by training 2D-FAN primarily on synthetic data, and there was a mismatch between training and testing landmark annotations.
	
	\section{Large Scale 3D Faces in-the-Wild dataset} \label{sec:large_scale}
	
	Motivated by the scarcity of 3D face alignment annotations and the remarkable performance of 2D-FAN, we opted to create a large scale 3D face alignment dataset by converting all existing 2D face alignment annotations to 3D. To this end, we trained a 2D-to-3D FAN as described in Subsection \ref{sec:2D-3D} and guided it using the predictions of 2D-FAN, creating 3D landmarks for: 300-W test set, 300-VW (both training and all 3 testing datasets), Menpo (the whole dataset).
	
	\begin{figure}[!htb]
		\begin{center}
			\centering
			\includegraphics[height=1.7in,trim={0cm 0cm 0cm 0cm},clip]{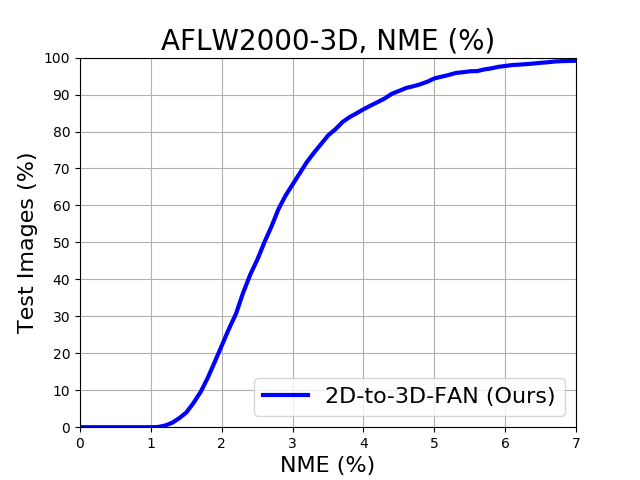}
		\end{center}
		\caption{NME on AFLW2000-3D, between the original annotations of \cite{zhu2016face} and the ones generated by 2D-to-3D-FAN. The error is mainly introduced by the automatic annotation process of \cite{zhu2016face}. See Fig.~\ref{fig:large_scale_aflw_2000} for visual examples.}
		\label{fig:results_aflw2Dto3D}
	\end{figure}
	
\begin{figure}[!htb]
		\begin{center}
			\centering
			\includegraphics[height=1.67in,trim={0.5cm 0.5cm 0.5cm 0.5cm},clip]{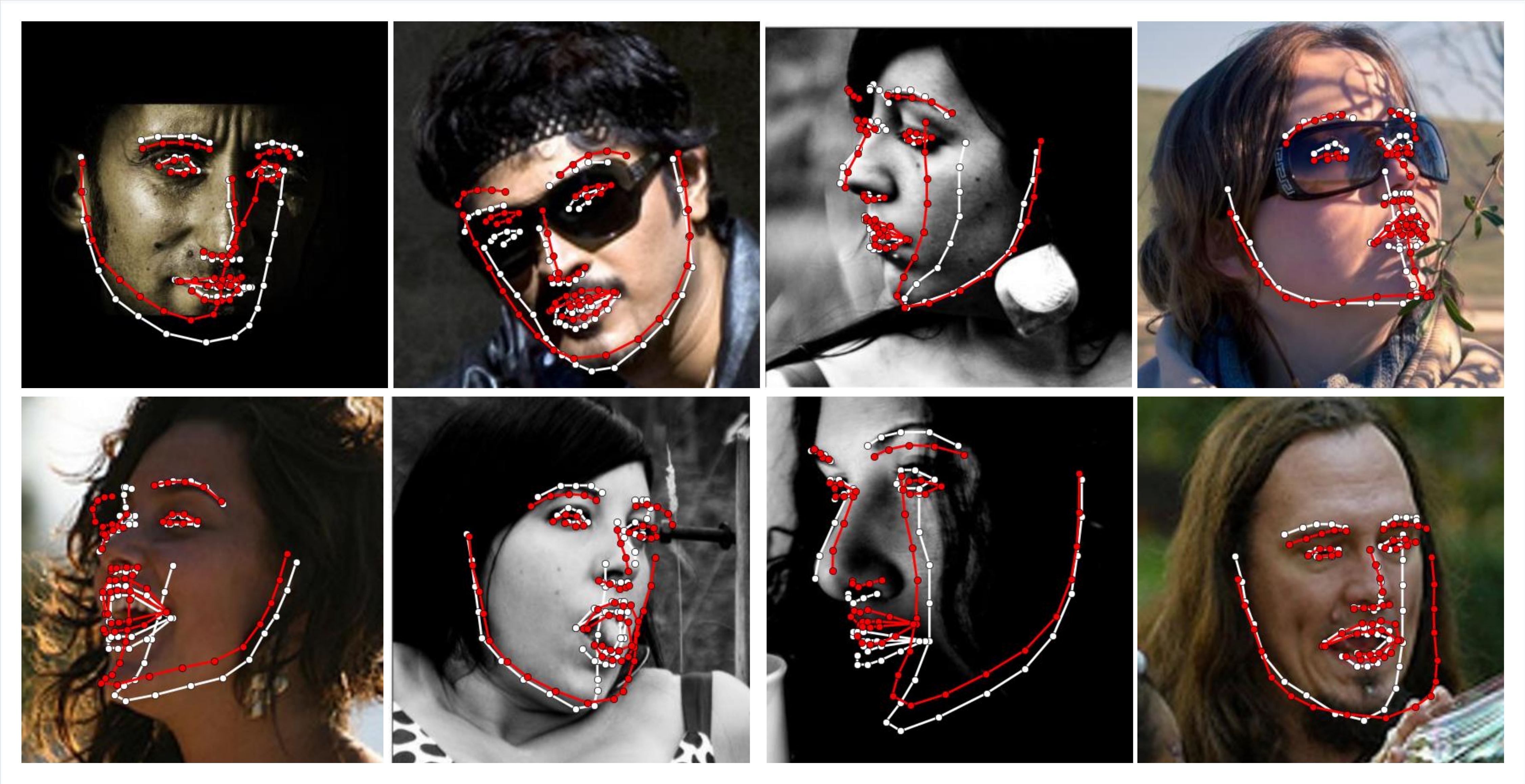}
		\end{center}
		\caption{Fittings with the highest error from AFLW2000-3D (NME 7-8\%). Red: ground truth from \cite{zhu2016face}. White: predictions of 2D-to-3D-FAN. In most cases, our predictions are more accurate than the ground truth.}
		\label{fig:large_scale_aflw_2000}
	\end{figure}

	Evaluating 2D-to-3D is difficult: the only available 3D face alignment in-the-wild dataset (not used for training) is AFLW2000-3D \cite{zhu2016face}. Hence, we applied our pipeline (consisting of applying 2D-FAN for producing the 2D landmarks and then 2D-to-3D FAN for converting them to 3D) on AFLW2000-3D and then calculated the error, shown in Fig.~\ref{fig:results_aflw2Dto3D} (note that for normalization purposes, 2D bounding box annotations are still used). The results show that there is discrepancy between our 3D landmarks and the ones provided by \cite{zhu2016face}. After removing a few failure cases (19 in total, which represent 0.9\% of the data), Fig.~\ref{fig:large_scale_aflw_2000} shows 8 images with the highest error between our 3D landmarks and the ones of \cite{zhu2016face}. It is evident, that this discrepancy is mainly caused from the semi-automatic annotation pipeline of \cite{zhu2016face} which does not produce accurate landmarks especially for images with difficult poses.  
	
	By additionally including AFLW2000-3D into the aforementioned datasets, overall, \texttildelow 230,000 images were annotated in terms of 3D landmarks leading to the creation of the Large Scale 3D Faces in-the-Wild dataset (LS3D-W), the largest 3D face alignment dataset to date.
	
	\section{3D face alignment} \label{sec:3D}
	
	This Section evaluates 3D-FAN trained on 300-W-LP-3D, on LS3D-W (described in the previous Section) i.e. on the 3D landmarks of the 300-W test set, 300-VW (both training and test sets), and Menpo (the whole dataset) and AFLW2000-3D (re-annotated). Overall, 3D-FAN is evaluated on \texttildelow 230,000 images. Note that compared to the 2D experiments reported in Section \ref{sec:2D}, more images in large poses have been used as our 3D experiments also include AFLW2000-3D and the profile images of Menpo (\texttildelow 2000 more images in total).  
	
\begin{figure*}[h]
		\vspace{-0.5cm}
		\centering
		\begin{subfigure}[t]{0.3\textwidth}
			\includegraphics[height=1.7in,trim={0cm 0cm 0cm 0cm},clip]{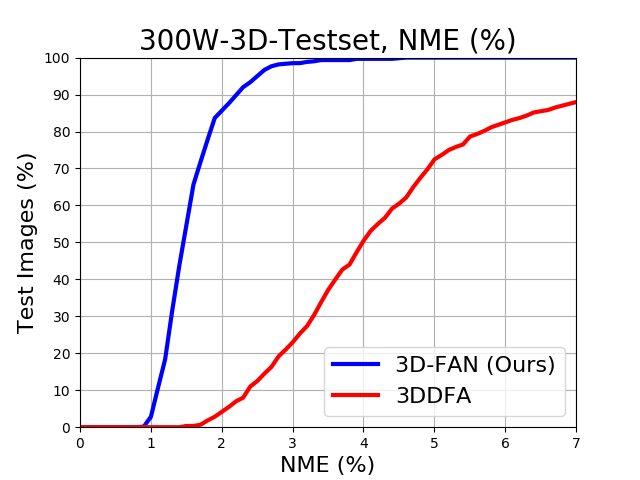}
			\caption{300-W-3D Test set}
			\label{fig:300W-3D_bb}
		\end{subfigure}
		~
		\begin{subfigure}[t]{0.3\textwidth}
			\includegraphics[height=1.7in,trim={0cm 0cm 0cm 0cm},clip]{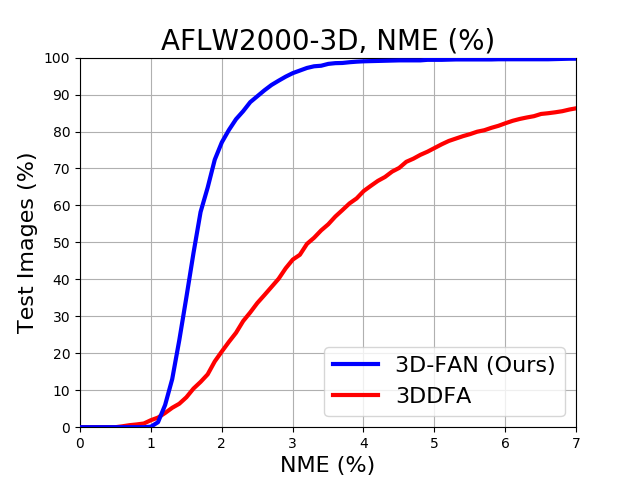}
			\caption{AFLW2000-3D, re-annotated using 2D-to-3D FAN.}
			\label{fig:AFLW2000-comp}
		\end{subfigure}
		~
		\begin{subfigure}[t]{0.3\textwidth}
			\includegraphics[height=1.7in,trim={0cm 0cm 0cm 0cm},clip]{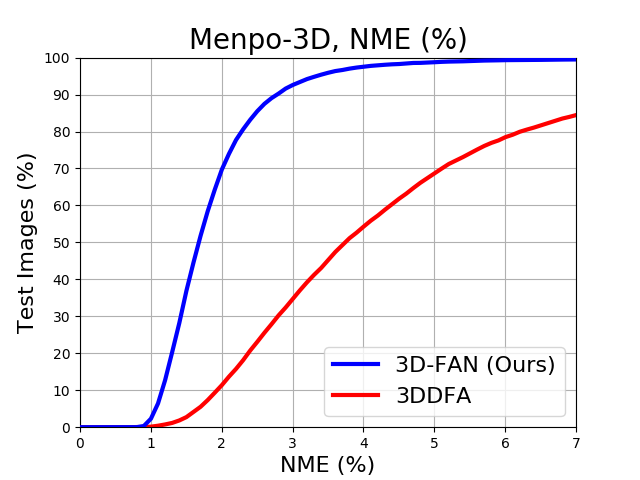}
			\caption{Menpo-3D}
			\label{fig:menpo-3D_bb}
		\end{subfigure}
		~
		\begin{subfigure}[t]{0.3\textwidth}
			\includegraphics[height=1.7in,trim={0cm 0cm 0cm 0cm},clip]{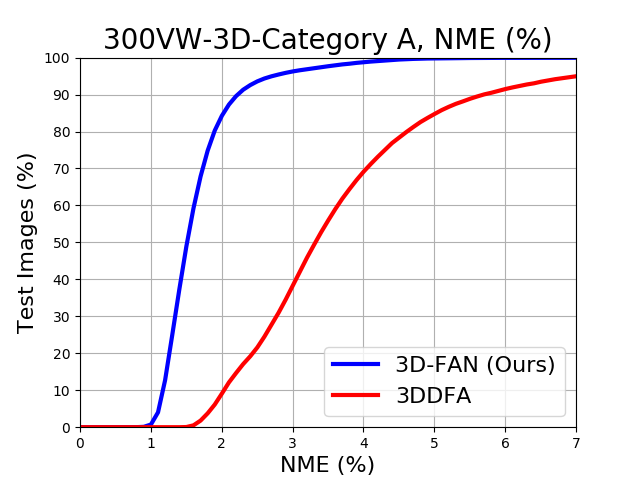}
			\caption{300VW-3D Category A}
			\label{fig:300VW-3D_A}
		\end{subfigure}
		~
		\begin{subfigure}[t]{0.3\textwidth}
			\includegraphics[height=1.7in,trim={0cm 0cm 0cm 0cm},clip]{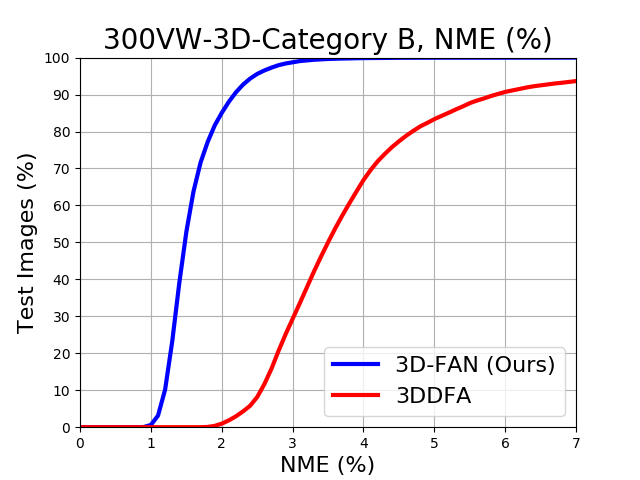}
			\caption{300VW-3D Category B}
			\label{fig:300VW-3D_B}
		\end{subfigure}
		~
		\begin{subfigure}[t]{0.3\textwidth}
			\includegraphics[height=1.7in,trim={0cm 0cm 0cm 0cm},clip]{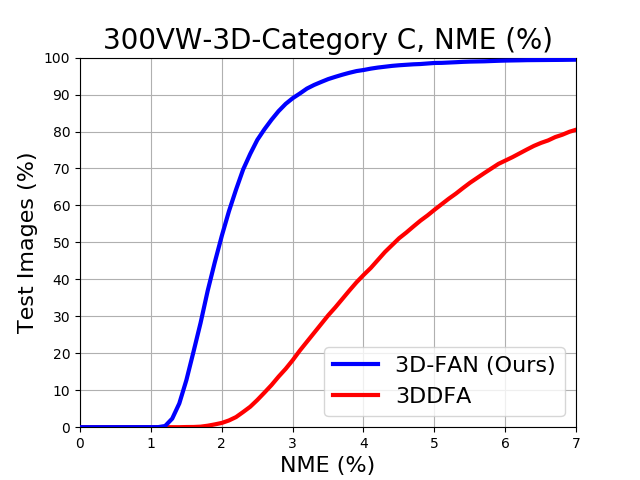}
			\caption{300VW-3D Category C}
			\label{fig:300VW-3D_C}
		\end{subfigure}
		\caption{3D face alignment experiments: NME (all 68 points used) on the newly introduced LS3D-W dataset.}
		\label{fig:300W-3D}
	\end{figure*}

	The results of our 3D face alignment experiments on 300-W test set, 300-VW, Menpo and AFLW2000-3D are shown in Fig.~\ref{fig:300W-3D}. We additionally report the performance of the state-of-the-art method of 3DDFA (trained on the same dataset as 3D-FAN) on all datasets. \newline \textbf{Conclusion:} 3D-FAN essentially produces the same accuracy on all datasets largely outperforming 3DDFA. This accuracy is slightly increased compared to the one achieved by 2D-FAN, especially for the part of the error curve for which the error is less than 2\% something which is not surprising as now the training and testing datasets are annotated using the same mark-up.
	
	\section{Ablation studies} \label{sec:AS}
	
	To further investigate the performance of 3D-FAN under challenging conditions, we firstly created 
	a dataset of 7,200 images from LS3D-W so that there is an equal number of images in yaw angles $[0^o-30^o]$,  $[30^o-60^o]$ and $[60^o-90^o]$. We call this dataset LS3D-W Balanced. Then, we conducted the following experiments:
	
	\begin{table}[!htbp]
		\small
		\begin{center}
			\begin{tabular}{|l|c|c|}
				\hline
				Yaw & \#images & 3D-FAN (Ours)  \\
				\hline\hline
				$[0^o-30^o]$ & 2400 &73.5\%\\
				$[30^o-60^o]$  & 2400 & 74.6\%\\
				$[60^o-90^o]$  & 2400 & 68.8\%\\
				\hline
			\end{tabular}
		\end{center}
		\caption{AUC (calculated for a threshold of 7\%) on the LS3D-W Balanced for different yaw angles.}
		\label{tab:ablation_yaw}
	\end{table}
	
	\textbf{Performance across pose.} We report the performance of 3D-FAN on LS3D-W Balanced for each pose separately in terms of the Area Under the Curve (AUC) (calculated for a threshold of 7\%) in Table~\ref{tab:ablation_yaw}. We observe only a slight degradation of performance for very large poses ($[60^o-90^o]$). We believe that this is to some extent to be expected as 3D-FAN was largely trained with synthetic data for these poses (300-W-LP-3D). This data was produced by warping frontal images (i.e. the ones of 300-W) to very large poses which causes face distortion especially for the face region close to the ears. \newline
	\textbf{Conclusion:} Facial pose is not a major issue for 3D-FAN. \newline \begin{figure}[!htb]
		\begin{center}
			\centering
			\includegraphics[height=1.7in,trim={0cm 0cm 0cm 0cm},clip]{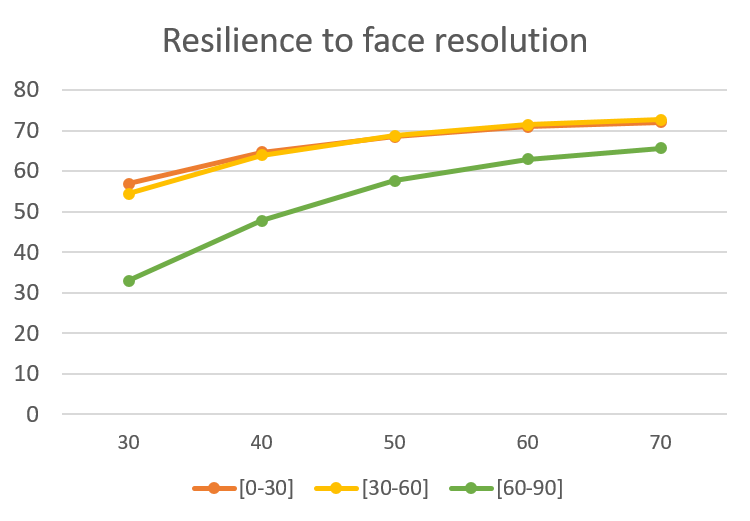}
		\end{center}
		\caption{AUC on the LS3D-W Balanced for different face resolutions. Up to 30px, performance remains high.}
		\label{fig:ablation_resolution}
	\end{figure}
	\textbf{Performance across resolution.} We repeated the previous experiment but for different face resolutions (resolution is reduced relative to the face size defined by the tight bounding box) and report the performance of 3D-FAN in terms of AUC in Fig.~\ref{fig:ablation_resolution}. Note that we did not retrain 3D-FAN to particularly work for such low resolutions. We observe significant performance drop for all poses only when the face size is as low as 30 pixels. \newline
	\textbf{Conclusion:} Resolution is not a major issue for 3D-FAN. 
	
	\begin{table}[!htbp]
		\small
		\begin{center}
			\begin{tabular}{|l|c|c|c|}
				\hline
				Noise & $[0^o-30^o]$ &  $[30^o-60^o]$  & $[60^o-90^o]$\\
				\hline\hline
				0\% &74.5\% &75.2\% & 69.8\%\\
				10\%  & 73.5\% &74.6\% & 68.8\%\\
				20\%  & 70.8\% &71.7\% & 66.1\%\\
				30\% & 63.8\% &63.5\% & 57.2\%\\
				\hline
			\end{tabular}
		\end{center}
		\caption{AUC on the LS3D-W Balanced for different levels of initialization noise. The network was trained with a noise level of up to 20\%.}
		\label{tab:ablation_init}
	\end{table}
	
	\textbf{Performance across noisy initializations.} For all reported results so far, we used 10\% of noise added to the ground truth bounding boxes. Note that 3D-FAN was trained with noise level of 20\%. Herein, we repeated the previous experiment but for different noise levels and report the performance of 3D-FAN in terms of AUC in Table~\ref{tab:ablation_init}. We observe only small performance decrease for noise level equal to 30\% which is greater than the level of noise that the network was trained with. \newline  
	\textbf{Conclusion:} Initialization is not a major issue for 3D-FAN.
	\begin{table}[!htbp]
		\small
		\begin{center}
			\begin{tabular}{|l|c|c|c|}
				\hline
				\#params & $[0^o-30^o]$ &  $[30^o-60^o]$  & $[60^o-90^o]$\\
				\hline\hline
				2M & 70.9\% &69.9\% & 55.8\%\\
				4M  & 71.0\% &70.5\% & 57.0\%\\
				6M  & 71.5\% &71.1\% & 58.3\%\\
				12M & 72.7\% &72.7\% & 67.1\%\\
				18M  & 73.4\% &74.2\% & 68.3\%\\
				24M  & 73.5\% &74.6\% & 68.8\%\\
				\hline
			\end{tabular}
		\end{center}
		\caption{AUC on the LS3D-W Balanced for various network sizes. Between 12-24M parameters, performance remains almost the same.}
		\label{tab:ablation_parameters}
	\end{table}
	\newline
	\textbf{Performance across different network sizes.} For all reported results so far, we used a very powerful 3D-FAN with 24M parameters. Herein, we repeated the previous experiment varying the number of network parameters and report the performance of 3D-FAN in terms of AUC in Table~\ref{tab:ablation_parameters}. The number of parameters is varied by firstly reducing the number of HG networks used from 4 to 1. Then, the number of parameters was dropped further by reducing the number of channels inside the building block. It is important to note that even then biggest network is able to run on 28-30 fps on a TitanX GPU while the smallest one can reach 150 fps. We observe that up to 12M, there is only a small performance drop and that the network's performance starts to drop significantly only when the number of parameters becomes as low as 6M. \newline  
	\textbf{Conclusion:} There is a moderate performance drop vs the number of parameters of 3D-FAN. We believe that this is an interesting direction for future work.
	
	\section{Conclusions}
	
	We constructed a state-of-the-art neural network for landmark localization, trained it for 2D and 3D face alignment, and evaluate it on hundreds of thousands of images. Our results show that our network nearly saturates these datasets, showing also remarkable resilience to pose, resolution, initialization, and even to the number of the network parameters used. Although some very unfamiliar poses were not explored in these datasets, there is no reason to believe, that given sufficient data, the network does not have the learning capacity to accommodate them, too.
	
	
	\section{Acknowledgments} Adrian Bulat was funded by a PhD scholarship from the University of Nottingham. This work was supported in part by the EPSRC project EP/M02153X/1 Facial Deformable Models of Animals.
	

	
    


\newpage
{\small
\bibliographystyle{ieee}
\bibliography{egbib}
}

\clearpage

\appendix
\renewcommand{\thesection}{A\arabic{section}}

\section{Additional numeric  results}
\begin{figure}[!htb]
	\begin{center}
		\centering
		\includegraphics[height=2.2in,trim={0cm 0cm 0cm 0cm},clip]{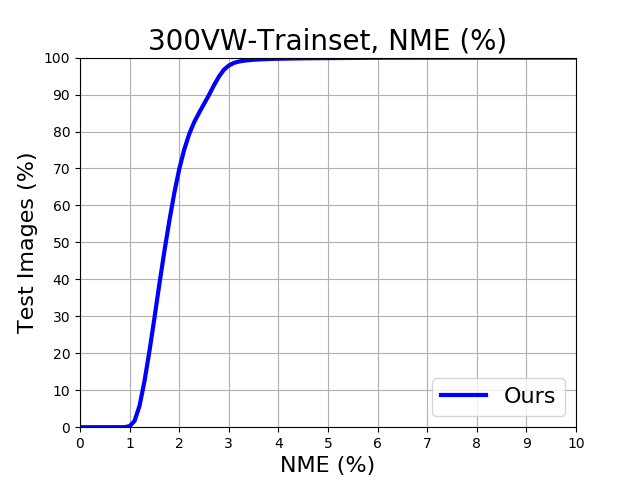}
	\end{center}
	\caption{NME (all 68 points used) on 300-VW training set.}
	\label{fig:300VW_trainset}
\end{figure}

Fig.~\ref{fig:300VW_trainset} provides additional numerical results for 2D-FAN on the 300-VW training set. 

\section{Additional visual results}    

\begin{figure*}[!htb]
	\begin{center}
		\centering
		\includegraphics[height=8.6in,trim={2cm 5cm 2cm 2.9cm},clip]{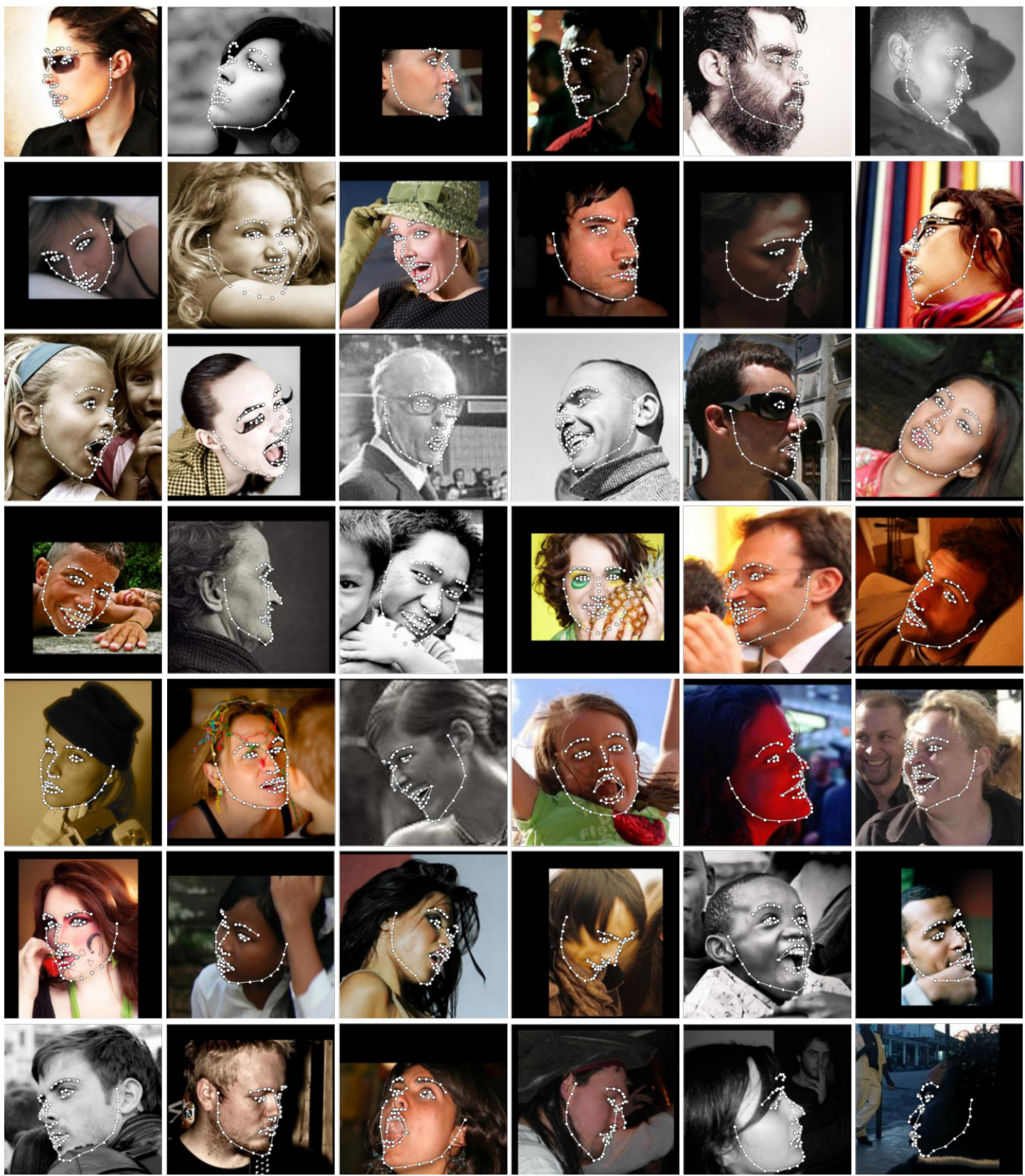}
	\end{center}
	\caption{Fitting examples produced by \textbf{2D-FAN} on LS3D-W balanced dataset.}
	\label{fig:examples_2DFAN}
\end{figure*}

\begin{figure*}[!htb]
	\begin{center}
		\centering
		\includegraphics[height=8.6in,trim={0cm 0.5cm 0cm 0cm},clip]{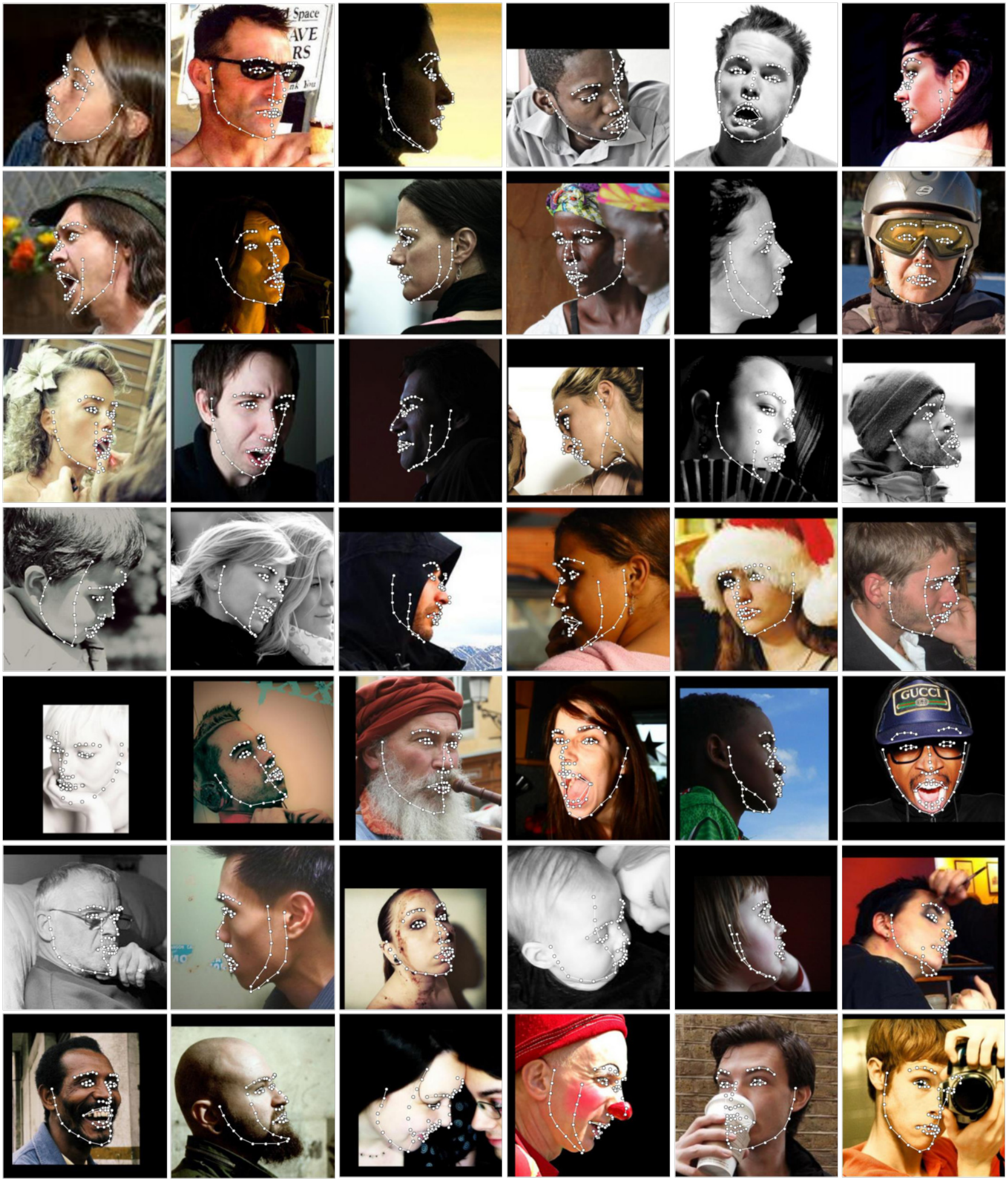}
	\end{center}
	\caption{Fitting examples produced by \textbf{3D-FAN} on LS3D-W balanced dataset.}
	\label{fig:examples_3DFAN}
\end{figure*}

Figs. \ref{fig:examples_2DFAN} and \ref{fig:examples_3DFAN} show a series of randomly picked-out samples produced by 2D-FAN and 3D-FAN, respectively, on LS3D-W balanced. Notice that our method copes very well with extreme poses, expressions, and lighting conditions.

\section{Full 3D face alignment}

In this Section, we present an extension of 2D-to-3D-FAN capable of additionally predicting the z coordinate of the facial landmarks.

\begin{figure}[!htb]
	\begin{center}
		\centering
		\includegraphics[height=1.1in,trim={0.5cm 0.5cm 0.5cm 0.5cm},clip]{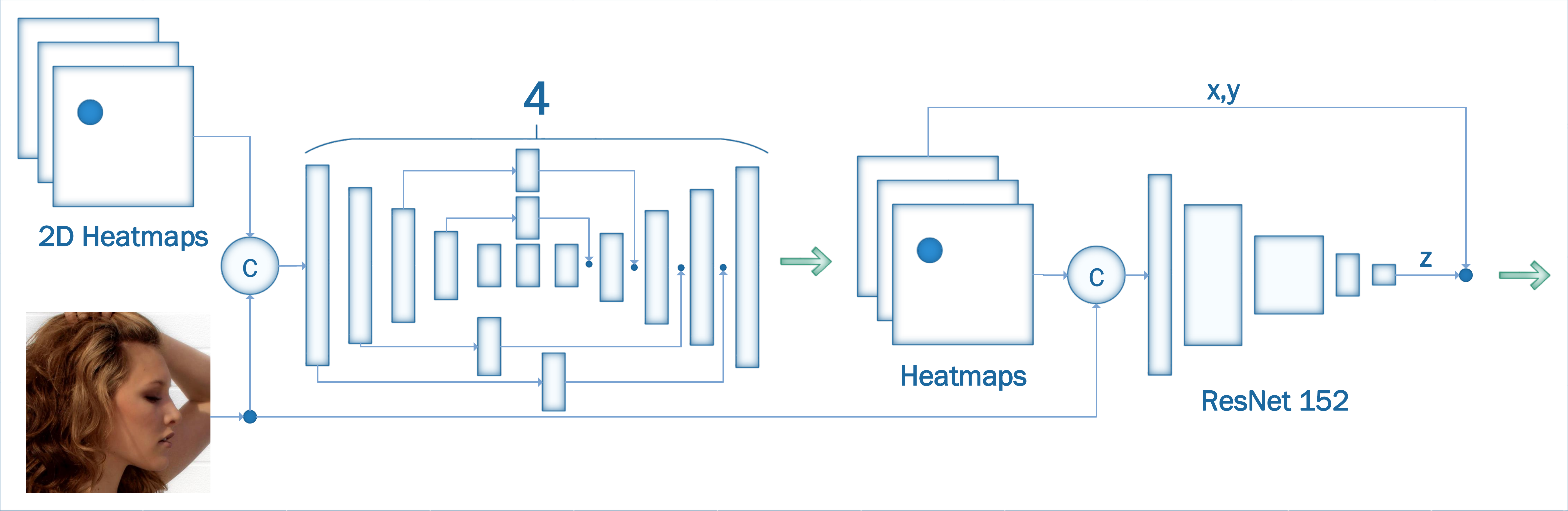}
	\end{center}
	\caption{The Full-2D-to-3D-FAN network used for the prediction of the $x,y,z$ coordinates. The network takes as input an RGB image and the 2D landmarks and outputs the corresponding 3D landmarks.}
	\label{fig:fan-3d-full}
\end{figure}

\begin{figure}[!htb]
	\begin{center}
		\centering
		\includegraphics[height=2.2in,trim={0cm 0cm 0cm 0cm},clip]{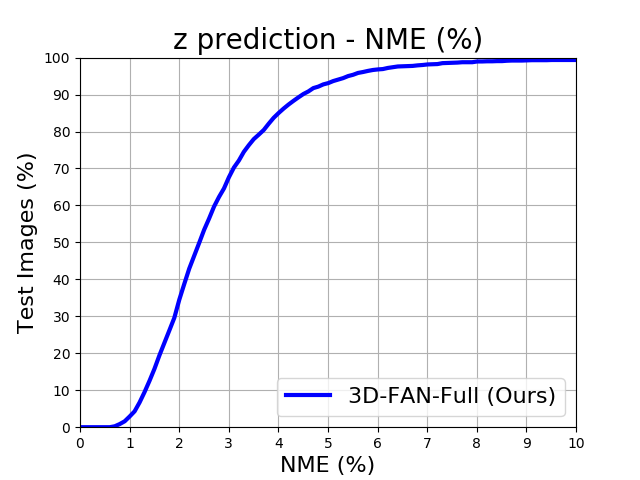}
	\end{center}
	\caption{NME on AFLW2000-3D, between the original annotations of \cite{zhu2016face} and the ones generated by 3D-FAN-Full for depth (z coordinate).}
	\label{fig:results_aflw3D_full}
\end{figure}

\begin{figure*}[!htb]
	\begin{center}
		\centering
		\includegraphics[height=7.6in,trim={0.5cm 0.5cm 0.5cm 0.5cm},clip]{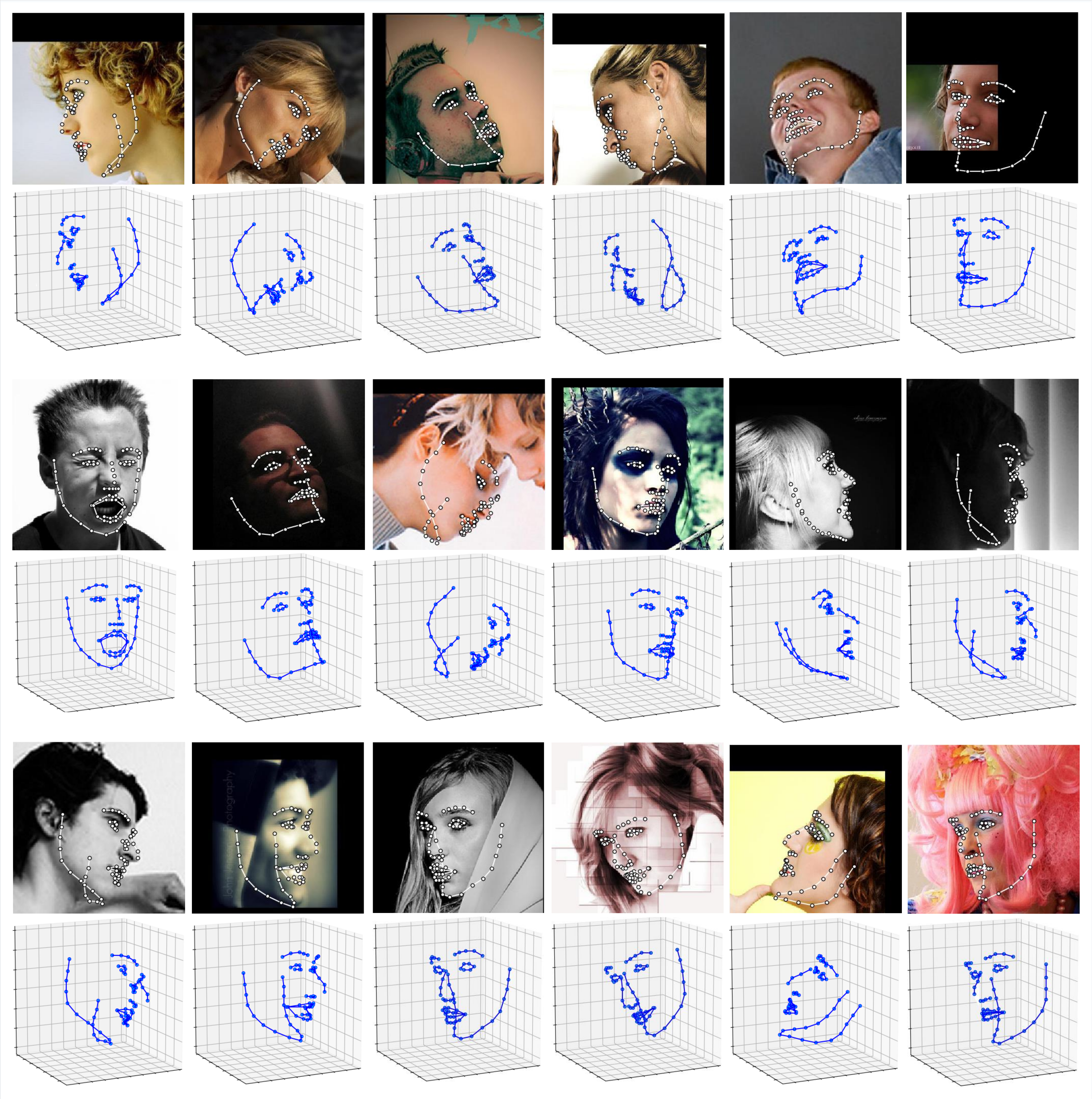}
	\end{center}
	\caption{Full 3D fitting examples produced by \textbf{Full-2D-to-3D-FAN} on AFLW2000-3D dataset.}
	\label{fig:examples_3Dfull}
\end{figure*}

Similarly to \cite{bulat2016two}, we construct Full-2D-to-3D-FAN by introducing a second subnetwork for estimating the $z$ coordinate (i.e. the depth of each keypoint) on top of 2D-to-3D-FAN. The input to the new subnetwork is the stacked heatmaps produced by 2D-to-3D-FAN alongside the RGB image. The heatmaps play a key role by showing the network where to ``look'' (i.e at which location should the depth be predicted) incorporating, at the same time, additional pose related information. The proposed subnetwork is based on a ResNet-152 {\cite{he2016deep}} adapted to accept $3+N$ input channels and to output a vector $N \times 1$ instead of $1000 \times 1$. The network was trained using the L2 loss for 50 epochs and the same learning rates used for the rest of the networks. Fig.~\ref{fig:results_aflw3D_full} reports the numerical error of Full-2D-to-3D-FAN on AFLW2000-3D. For visual results, see Fig.~\ref{fig:examples_3Dfull}.

\end{document}